\pdfoutput=1

\documentclass[11pt]{article}
\usepackage{mdframed}
\usepackage{fancyvrb}
\usepackage{listings}

\usepackage{array}
\usepackage{booktabs}
\usepackage{adjustbox}
\usepackage{float}

\usepackage{subcaption}
\usepackage{graphicx}
\usepackage{tabularx}
\usepackage{multirow}

\usepackage[preprint]{acl}

\usepackage{times}
\usepackage{latexsym}

\usepackage[T1]{fontenc}

\usepackage[utf8]{inputenc}

\usepackage{microtype}

\usepackage{inconsolata}

\usepackage{graphicx}
\usepackage{amssymb}
\usepackage{amsmath}
\usepackage{booktabs}
\usepackage{enumitem}

\usepackage{soul}
\usepackage{xcolor}
\definecolor{customgreen}{rgb}{0.5, 0.8, 0.5} 
\sethlcolor{customgreen!80} 
\definecolor{lightgreen}{rgb}{0.5, 1.0, 0.5} 

\definecolor{LightGreen}{rgb}{0.5, 0.9, 0.5}
\definecolor{MediumGreen}{rgb}{0.5, 0.8, 0.5}
\definecolor{DarkGreen}{rgb}{0.0, 0.5, 0.0}

\definecolor{myblue}{RGB}{31, 174, 235}
\definecolor{mypink}{RGB}{251, 155, 156}
\definecolor{light_pink}{HTML}{ffd5d5}
\usepackage[most]{tcolorbox}
\tcbset{on line,
        boxsep=1pt, left=0pt,right=0pt,top=0pt,bottom=0pt,
        colframe=white,colback=light_pink,
        highlight math style={enhanced}
        }
\newcommand{\ff}{f}
\newcommand{\vv}{\mathbf{v}_{\ff}}
\newcommand{\ww}{\mathbf{w}}

\newcommand{\maxact}[0]{\texttt{MaxAct}}
\newcommand{\vocabproj}[0]{\texttt{VocabProj}}
\newcommand{\tokenshift}[0]{\texttt{TokenChange}}
\newcommand{\rawnsemble}[0]{\texttt{Ensemble Raw}}
\newcommand{\rawnsembles}[1]{\texttt{Ensemble Raw (#1)}}
\newcommand{\catsemble}[0]{\texttt{Ensemble Concat}}

\newcommand{\srawnsembles}[1]{\texttt{EnsembleR (#1)}}

\newcommand{\scatsembles}[1]{\texttt{EnsembleC (#1)}}

\newcommand{\gemma}{Gemma-2}
\newcommand{\gemmaF}{Gemma-2 2B}
\newcommand{\gemmaNF}{Gemma-2 9B}
\newcommand{\gemmaTF}{Gemma-2 27B}
\newcommand{\llama}{Llama-3.1}
\newcommand{\llamaF}{Llama-3.1 8B}
\newcommand{\gpt}{GPT-2 small}
\newcommand{\llamaI}{Llama-3.1 Instruct}
\newcommand{\llamaIF}{Llama-3.1 8B Instruct}
\newcommand{\gptmini}{GPT-4o mini}
\newcommand{\gemini}{Gemini 1.5 Pro}
\newcommand{\gemmasae}{Gemma Scope}
\newcommand{\llamasae}{Llama Scope}
\newcommand{\gptsae}{OpenAI SAE}
\newcommand{\pedia}{Neuronpedia}

\title{Enhancing Automated Interpretability \\ with Output-Centric Feature Descriptions}

\author{
 \vspace{5px}
Yoav Gur-Arieh$^1$ ~~~ Roy Mayan$^{1}$\thanks{\ Equal contribution} ~~~ Chen Agassy$^{1*}$ ~~~ Atticus Geiger$^2$ ~~~ Mor Geva$^1$ \\ \vspace{3px}
$^1$Blavatnik School of Computer Science and AI, Tel Aviv University\\
$^2$Pr(Ai)$^2$R Group\\
\small{\texttt{\{yoavgurarieh@mail,roymayan@mail,chenagassy@mail,morgeva@tauex\}.tau.ac.il}},
\small{\texttt{atticusg@gmail.com}}
}

\begin{document}

\maketitle

\begin{abstract}

Automated interpretability pipelines generate natural language descriptions for the concepts represented by features in large language models (LLMs), such as \textit{plants} or \textit{the first word in a sentence}. These descriptions are derived using \textit{inputs} that activate the feature, which may be a dimension or a direction in the model's representation space. However, identifying activating inputs is costly, and the mechanistic role of a feature in model behavior is determined both by how inputs cause a feature to activate and by how feature activation affects \textit{outputs}. Using steering evaluations, we reveal that current pipelines provide descriptions that fail to capture the causal effect of the feature on outputs. To fix this, we propose efficient, output-centric methods for automatically generating feature descriptions. These methods use the tokens weighted higher after feature stimulation or the highest weight tokens after applying the vocabulary ``unembedding'' head directly to the feature. Our output-centric descriptions better capture the causal effect of a feature on model outputs than input-centric descriptions, but combining the two leads to the best performance on both input and output evaluations. Lastly, we show that output-centric descriptions can be used to find inputs that activate features previously thought to be ``dead''.

\end{abstract}
\section{Introduction}
\label{sec:introduction}

\begin{figure}[t]
    \centering
    \includegraphics[width=0.99\linewidth]{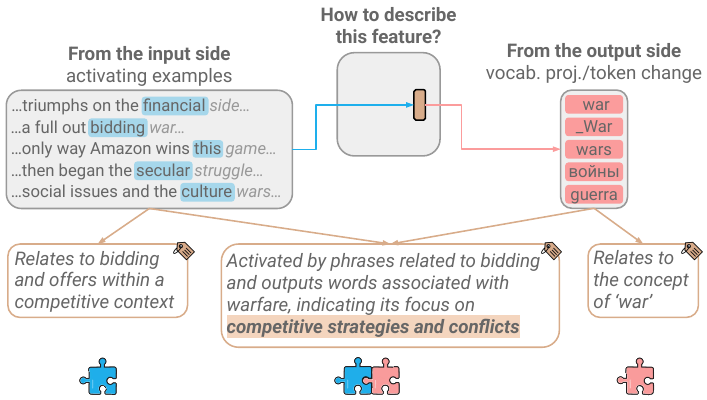}
    \caption{We posit that a faithful description of a feature should consider both model inputs that activate it (\textbf{\textcolor{myblue}{left}}, marked words cause the highest activations) and the effect it introduces to the model's outputs (\textbf{\textcolor{mypink}{right}}).}
    \label{fig:intro}
\end{figure}

Understanding how language models represent concepts in a real-valued vector space has long been a central challenge in NLP \cite{mikolov2013,karpathy2015visualizing, bau2018identifying, NEURIPS2020_c74956ff, dai-etal-2022-knowledge, park2024geometrycategoricalhierarchicalconcepts}. 
Recent efforts to scale this process use automated interpretability pipelines, where large language models (LLMs) describe the concepts encoded by features, i.e., small model components such as neurons or directions in activation space, \textit{based on inputs that activate them} \cite{bills2023language, bricken2023monosemanticity,paulo2024automaticallyinterpretingmillionsfeatures, choi2024automatic}.
However, despite its wide adoption, solely relying on the inputs activating a feature to describe it has practical limitations and theoretical pitfalls.

First, given the large corpora modern LLMs are trained on, obtaining these examples can be costly and nearly impossible in cases when features are described by data instances that are not publicly available. This practical limitation increases the compute and data needed for automated interpretability.
Second, the concept represented by a feature is determined by the causal role of that feature in model behavior, namely, how model inputs cause the feature to activate and how a feature causes model outputs to change \cite{mueller2024}. Using only inputs to characterize a feature is ungrounded in the causal mechanisms driving model behavior, which introduces pitfalls. For example, different datasets can lead to inconsistent feature descriptions \cite{bolukbasi2021interpretabilityillusionbert} or to classifying features as ``dead'' due to lack of activation \cite{gao2024scalingevaluatingsparseautoencoders, templeton2024scaling}. 
Last, a common use of feature descriptions is controlling model behavior through ``steering'', i.e., stimulating a feature to control the model's outputs \cite{8100128, li2023inferencetime, rimsky-etal-2024-steering, templeton2024scaling, obrien2024steeringlanguagemodelrefusal}. Therefore, good feature descriptions for steering should be output-centric.

To overcome these limitations, we propose two output-centric methods for enhancing automated interpretability pipelines (see Figure~\ref{fig:intro} for illustration). 
The first method, called \vocabproj{}, uses the prominent tokens in the projection of a feature to the model's vocabulary space \cite{geva-etal-2022-transformer, bloom2024understanding}. 
The second method, called \tokenshift{}, considers the tokens whose probabilities in the model's output distribution change the most when the feature is amplified.
Notably, these methods are substantially more computationally efficient than generating descriptions based on activating inputs; \vocabproj{} requires a single matrix multiplication, and \tokenshift{} involves running the model on a few inputs.

We compare the descriptions generated by these methods with those generated based on maximum activating inputs (dubbed \maxact{}) using two evaluations: \textit{input-based} and \textit{output-based} (see Figure~\ref{fig:evals} for illustration). The input-based evaluation assesses how accurately a description identifies what triggers the feature, whereas the output-based evaluation measures how effectively the description captures the causal impact of the feature's activation on the model's output.

Experiments over neuron-aligned and sparse autoencoder (SAE) features from both the residual and MLP layers of multiple LLMs reveal substantial differences between the methods and the descriptions they yield.
While \maxact{} typically outperforms \vocabproj{} and \tokenshift{} on the input-based evaluation, it is generally worse in capturing the feature's effect on the model's generation. Moreover, the gap between \maxact{} and \vocabproj{} in describing the inputs activating a given feature is sometimes small, suggesting that the latter can serve as a cheap replacement in such cases.
Last, ensembles of the three methods consistently achieve the best performance across both evaluations, providing strong empirical evidence for the benefits of incorporating output-centric methods into automated interpretability pipelines.

Further analysis sheds light on those benefits. We observe that descriptions generated by output-centric methods are often abstractions of their input-centric counterparts, and that the composition of the input- and output-centric descriptions of a feature can in some cases provide a new meaning (e.g. Figure~\ref{fig:intro}). Additionally, experiments with Gemma-2 SAEs show that output-centric methods can be used to efficiently discover inputs that activate ``dead'' features, for which no activating inputs had previously been identified.

To summarize, our work makes the following contributions: (a) we propose a two-faceted evaluation framework for feature descriptions, examining them through complementary input and output lenses (b) we highlight key drawbacks of using \maxact{}, the common method used today in automated interpretability pipelines, to obtain feature descriptions in LLMs, (c) we propose output-centric methods to mitigate these limitations, (d) our experiments demonstrate the effectiveness of each approach and that their combination yields more faithful feature descriptions, (e) our analysis provides insights into the benefits in combining input- and output-centric methods.
By producing more faithful and complete feature descriptions, our approach can enhance downstream applications such as model editing, machine unlearning, and circuit analysis \citep[e.g.,][]{wu-etal-2023-depn, farrell2024applyingsparseautoencodersunlearn, marks2025sparse}.
We release our code and generated feature descriptions at \url{https://github.com/yoavgur/Feature-Descriptions}.

\section{Problem Setup}
\label{sec:problem}

We focus on the problem of automatically describing atomic units of computation in LLMs called \textit{features}. As the exact nature of features is a hotly debated topic, we adopt the general framework of \citet{geiger2024causalabstractiontheoreticalfoundation} which we limit to real-valued features. Let $\mathcal{M}$ be our target LLM. Any hidden vector $\mathbf{v} \in \mathbb{R}^d$ in $\mathcal{M}$ can be transformed with an invertible \textit{featurizer} $\mathcal{F}: \mathbb{R}^d \to \mathbb{R}^k$ that maps the vector into a space of $k$ features. A single feature $\ff \in \mathbb{R}^k$ is simply a one-hot encoding which can be vectorized using $\vv = \mathcal{F}^{-1}(\ff)$.
This framework supports a variety of features, including neurons (axis-aligned dimensions) in MLPs \cite{geva-etal-2022-transformer}, sets of orthogonal directions \cite{geigerDAS, huang-etal-2024-ravel, park2024linearrepresentationhypothesisgeometry}, sparse linear features from SAEs \cite{bricken2023monosemanticity, templeton2024scaling, huben2024sparse}, or even non-linear features, e.g. ``onion'' representations with a magnitude-based features \cite{csordas-etal-2024-recurrent}. 
 
During inference, the LLM constructs the vector $\mathbf{v}$ from the input, which can then be passed through $\mathcal{F}$ to determine the activation for each feature $\mathcal{F}(\mathbf{v})$. The possible values for activations are a result of the feature space, e.g. SAE features produced with a ReLU only have positive activations.

In this work, we consider the problem of 
automatically labeling the concept represented by a feature $\ff$. Namely, producing a human-understandable description text $s_{\ff}$ of the feature $\ff$.
Importantly, we want the method producing $s_{\ff}$ to be scalable, i.e. automatic and efficient, such that it can be integrated into large-scale pipelines that interpret millions of features in LLMs. This additional requirement excludes approaches that rely, for example, on manual human labeling.

A key question that arises is how to evaluate whether a description faithfully describes its corresponding feature. Here we observe that describing a feature is practically \textit{a two-faceted problem}; one can describe what inputs activate the feature, i.e. what inputs yield high feature activations, but they can also describe what this feature promotes in the model's output. Consider for example the feature illustrated in Figure~\ref{fig:intro}. The input side indicates that the feature activates mainly on competitive financial and business related sentences. Conversely, the output side shows that the feature amplifies the concept of war when activated. Only when considering the two sides together we see that the feature promotes the concept of war in social and business related scenarios, e.g., \textit{trade war}, \textit{bidding war}, and \textit{culture war}.
Notably, this formulation was also discussed in prior works; \citet{geva-etal-2021-transformer,geva-etal-2022-lm,geva-etal-2022-transformer} characterized MLP as key-value memories that promote specific concepts, and  \citet{antverg2022on, huang-etal-2023-rigorously} contended the importance of differentiating between the information encoded by the feature versus used by the model.

Despite the dual nature of this problem, existing automated interpretability pipelines \citep[e.g.,][]{bills2023language,paulo2024automaticallyinterpretingmillionsfeatures, choi2024automatic} have focused on one side of the problem. 
Namely, describing the inputs that activate the feature, while disregarding the feature's influence on the model's output.
For example, \citet{huang-etal-2023-rigorously} showed that neurons interpreted by \citet{bills2023language} lack causal influence on the concepts expressed in their generated descriptions. 
Therefore, we offer a more holistic approach, accounting for both the input and output of the model.

\section{Evaluation of Feature Descriptions}
\label{sec:scoring}

\begin{figure}[t]
    \centering
    \includegraphics[width=0.99\linewidth]{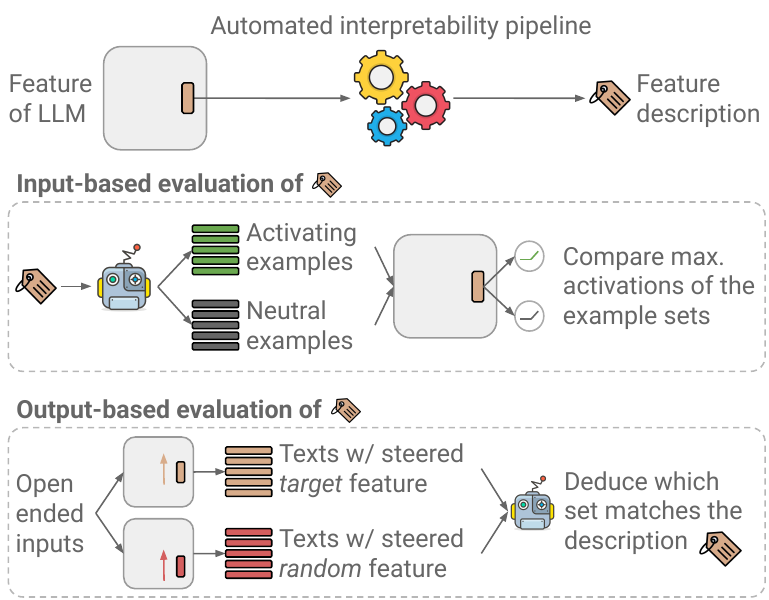}
    \caption{Illustration of our feature description evaluation, considering the description's faithfulness with respect to both the input (middle panel) and output (lower panel) of the model.}
    \label{fig:evals}
\end{figure}

We propose to evaluate how faithful a description is to its corresponding feature with the following complementary metrics, illustrated in Figure~\ref{fig:evals}.

\paragraph{Input-based Evaluation}

Following \citet{huang-etal-2023-rigorously, caden2024open}, we evaluate how well the description captures the inputs triggering the feature. Given a feature $\ff$, we feed its description $s_{\ff}$ generated by some method into an LLM, which is tasked to generate two sets of $k$ examples each: \textit{activating} and \textit{neutral}. These examples are expected and not expected to activate $\ff$ according to $s_{\ff}$, respectively (see  \S\ref{apx:evals} for examples and details regarding prompts). We then pass the generated examples through $\mathcal{M}$ and obtain $\ff$'s activation for each example, calculated as the max activation over all token positions in that example. We take the max over all token positions since it’s reasonable to expect $\ff$ to be activated highly even for just a single token, and not at all for the rest, following prior work that treats strong localized activation as meaningful \cite{bills2023language, choi2024automatic, paulo2024automaticallyinterpretingmillionsfeatures, voita-etal-2024-neurons}.
Let $\bar{m}_{\text{activating}}$ and $\bar{m}_{\text{neutral}}$ be the mean activations obtained for the activating and neutral examples, respectively.
The description $s_{\ff}$ is considered faithful if the mean activation for the activating examples exceeds that of the neutral examples, namely:
\begin{equation*}
    \bar{m}_{\text{activating}} >  \bar{m}_{\text{neutral}}
\end{equation*}
This evaluation is similar to those implemented in existing automated pipelines, which essentially measure \textit{how accurately the description captures the inputs that activate the feature}.

\paragraph{Output-based Evaluation}
To assess how faithful $s_{\ff}$ is with respect to $\ff$'s influence on the model's outputs, we evaluate $s_{\ff}$ against outputs generated by $\mathcal{M}$ when steering $\ff$ versus when steering another feature $\ff'$.
Concretely, we feed $\mathcal{M}$ open-ended prompts, such as \texttt{``<BOS> I think''} \cite{chalnev2024improvingsteeringvectorstargeting}, and let the model generate $n$ tokens three times -- one time while amplifying $\ff$ and two other times by amplifying two different random features $\ff'$ and $\ff''$. Amplification of a feature is done by clamping its activation to a high value $m$ \cite{templeton2024scaling}. 
Since finding an effective yet not destructive amplification level is challenging \cite{bhalla2024unifyinginterpretabilitycontrolevaluation, templeton2024scaling}, we run each input with varying levels of amplification while fixing the KL-divergence between the outputs of the steered model and the non-steered model \cite{paulo2024automaticallyinterpretingmillionsfeatures}, as calculated on a single next token prediction, averaged over all open ended prompts. 
This way we generate three sets of texts $\mathcal{T}_{\ff}$, $\mathcal{T}_{\ff'}$ and $\mathcal{T}_{\ff''}$. 
Next, we feed $s_{\ff}$ concatenated with $\mathcal{T}_{\ff}$, $\mathcal{T}_{\ff'}$ and $\mathcal{T}_{\ff''}$ to a judge LLM (see justification in \S\ref{apx:human_eval}), and task it to indicate which of the three sets matches $s_{\ff}$. The description $s_{\ff}$ is faithful if the LLM selects $\mathcal{T}_{\ff}$. Namely, we evaluate \textit{how well the description captures the feature's impact on the model's output}. For details, example generations and prompts used, see \S\ref{apx:evals}.

\section{Interpretability Methods}
\label{sec:generation}

We describe the methods used for automatically describing features in LLMs. These include the input-centric method prevalent today, two output-centric methods that describe a feature $\ff$ using its corresponding vector $\vv$, and their ensembles.

\paragraph{Max Activating Examples (\maxact{})}
Using the inputs that maximally activate a given feature to understand its function has been used extensively \cite{DBLP:journals/corr/abs-1812-09355, na2018discovery, bolukbasi2021interpretabilityillusionbert}. More recently, this method has been widely adopted and refined for automatically interpreting features at scale \cite{bills2023language, bricken2023monosemanticity, paulo2024automaticallyinterpretingmillionsfeatures, choi2024automatic, he2024llamascopeextractingmillions, huben2024sparse}. The method involves collecting feature activations in $\mathcal{M}$ across a large dataset. For each feature, $k$ examples are sampled from the dataset, prioritizing those with the highest activations, along with some examples from other activation quantiles \cite{bricken2023monosemanticity}. These examples are then fed to an explainer model, which is tasked with generating a description of the feature by the examples that activate it.

\paragraph{Vocabulary Projection (\vocabproj{})}
Building on \citet{geva-etal-2021-transformer,geva-etal-2022-lm,geva-etal-2022-transformer}, we propose to view the feature $\ff$ as an update to the model's output distribution. To interpret $\ff$'s contribution, we compute the feature vector $\mathcal{F}^{-1}(\ff) = \vv \in \mathbb{R}^d$ and project it to the vocabulary space to obtain a vector of logits $\ww \in \mathbb{R}^{|\mathcal{V}|}$ such that: 
\begin{equation*}
\ww = W_U \texttt{LayerNorm}(\vv)
\end{equation*}
where $\mathcal{V}$ is $\mathcal{M}$'s vocabulary, $\texttt{LayerNorm}$ is the final layer norm, and $W_U \in \mathbb{R}^{|\mathcal{V}| \times d}$ is the model's unembedding matrix.
We then examine the tokens corresponding to the top- and bottom-scoring entries in $\ww$, interpreting them as the tokens most promoted or suppressed, respectively. These tokens are then fed to an explainer model that generates a description for the feature.
For more details and other variants of this method, see \S\ref{subsec:vocabproj_variants}.

\paragraph{Token Change (\tokenshift{})}
This method describes the tokens whose logits in the model's output were most affected by amplifying the feature. Specifically, we pass $k$ random prompts sampled from some dataset through the model and collect the output logit values for each token position. Next, the feature is clamped to activation value $m$, and we collect the new logit values \cite{templeton2024scaling}. We then calculate the mean change in logit value per token across all positions and prompts. The list of tokens most affected by amplifying the feature is provided to an explainer model, which generates a description for the feature.

While both \vocabproj{} and \tokenshift{} are output-centric methods, \vocabproj{} is correlative and \tokenshift{} causally intervenes in the model's generation.

\paragraph{Ensembles}
To capture both the input and output sides of a feature, we propose combining the above approaches in two ways: (a) \rawnsemble{}: the raw data used by the methods is concatenated and fed to the explainer model. For example, in \rawnsembles{\maxact{}+\vocabproj{}} we would feed the explainer model the activating examples and top tokens in the vocabulary projection. (b) \catsemble{}: the description is simply a concatenation of the descriptions generated by the methods.
We also attempted to summarize the descriptions by the different methods with an LLM to produce a more cohesive description, but these ensembles performed worse across the board.

\section{Experiments}
\label{sec:exp}

In this section, we evaluate the above methods on our input- and output-based evaluations. Additional human evaluations are reported in \S\ref{apx:human_eval}.

\subsection{Experimental Setting}

\paragraph{Features} 
We analyze both features learned through SAEs and neurons in MLP layers, covering four LLMs of different sizes and families: \gemmaF{} \cite{team2024gemma}, \llamaF{} and \llamaIF{} \cite{dubey2024llama}, and \gpt{} \cite{radford2019language}. 
For \gemma{}, \llama{} and \gpt{}, we evaluate descriptions of SAE features trained on residual stream and MLP layers: \gemmasae{} 16K and 65K \cite{lieberum-etal-2024-gemma}, \llamasae{} 32K \cite{he2024llama}, and \gptsae{} 32K and 128K \cite{gao2024scalingevaluatingsparseautoencoders}. The activation function used by \gemmasae{} is JumpReLU \cite{rajamanoharan2024jumpingaheadimprovingreconstruction}, while both \llamasae{} and \gptsae{} use TopK-ReLU \cite{makhzani2014ksparseautoencoders}.
We randomly sample $n=40$ features per layer from every SAE, resulting in a total of 4,160 features for \gemma{}, 2,560 for \llama{} and 2,880 for \gpt{}.
For \llamaI{} we inspect a sample of $n=80$ MLP features per layer, with 2,560 features in total.

\paragraph{Description Generation}
We use the methods described in \S\ref{sec:generation} and generate descriptions for each feature, using \gptmini{} \cite{hurst2024gpt} as our explainer model to ensure consistency with descriptions from \pedia{} \cite{neuronpedia} and Transluce \cite{choi2024automatic}. 
For \maxact{}, we utilize the publicly available feature descriptions from these repositories. 
To validate these descriptions are comparable to those generated by us, we sampled 1,080 features and found their descriptions match those we generate for \maxact{} (see \S\ref{subsec:recreating_pedia}).

When generating ensembles from raw data (\rawnsemble{}), we rely on feature activation data from these same sources, using the top five activating sentences to keep in line with existing methods. Notably, Transluce generated descriptions for \llamaIF{} through a more complex process than \maxact{} \cite{choi2024automatic}, creating multiple descriptions from activating examples and selecting the best one using simulation scoring \cite{bills2023language}. For clarity, we refer to this method as \maxact{}++ and generate the \maxact{} descriptions for \llamaIF{} ourselves using the feature activation data from Transluce.

For \vocabproj{} and \tokenshift{}, we pass the top and bottom $t$ tokens to the explainer model \gptmini{} (see prompts in \S\ref{subsec:description_generation}). We set $t=50$ for \vocabproj{} and $t=20$ for \tokenshift{}. For \tokenshift{} we use $k=32$ random prompts of 32 tokens each from The Pile \cite{gao2020pile800gbdatasetdiverse}.

\paragraph{Description Evaluation}
For the input-based evaluation, we instruct \gemini{} \cite{team2024gemini} to generate five activating and five neutral sentences with respect to a given feature description. 
For the output-based evaluation, we prompt the model with three open-ended prompts, letting it generate up to 25 tokens while clamping the feature's activation value to $m$ for all token positions. For each prompt, we run the model four times with increasing clamping values, making the generations progressively more affected by the feature's output. This process results in 12 text generations for each of the sets $\mathcal{T}_{\vv}$, $\mathcal{T}_{\vv'}$, and $\mathcal{T}_{\vv''}$, which we provide to \gptmini{} \cite{hurst2024gpt} as a judge (see \S\ref{apx:evals} for more details and exact prompts). We select this model to minimize costs, given the lengthy prompts induced by the text sets.

\begin{table*}
\setlength{\belowcaptionskip}{-5pt}
\setlength{\tabcolsep}{3.5pt}
\centering
\footnotesize
\begin{tabular}{lcccccccc}
\toprule
      & \multicolumn{2}{c}{\textbf{\gemma{} Res. SAE}}  & \multicolumn{2}{c}{\textbf{\gemma{} MLP SAE}} & \multicolumn{2}{c}{\textbf{\llama{} Res. SAE}} & \multicolumn{2}{c}{\textbf{\llama{} Inst. MLP}} \\
     & Input & Output & Input & Output & Input & Output & Input & Output \\ \midrule
     \maxact{} & 56.6 $\pm$ 2.2 & 49.2 $\pm$ 2.2 & 50.4 $\pm$ 2.2 & 35.1 $\pm$ 2.1 & 30.3 $\pm$ 2.7 & 71.8 $\pm$ 2.6 & 85.6 $\pm$ 1.4 & 36.9 $\pm$ 1.9 \\
     \maxact{}++ & ~~- & ~~- & ~~- & ~~- & ~~- & ~~- & \textbf{89.8} $\pm$ 1.2 & ~~~39 $\pm$ 1.9 \\
     \vocabproj{} & 50.1 $\pm$ 2.2 & 56.5 $\pm$ 2.2 & 20.9 $\pm$ 1.8 & 37.2 $\pm$ 2.1 & 18.2 $\pm$ 2.2 & 64.2 $\pm$ 2.8 & 71.2 $\pm$ 1.8 & \textbf{45.8} $\pm$ 1.9 \\
     \tokenshift{} & 44.7 $\pm$ 2.2 & 54.9 $\pm$ 2.2 & 22.3 $\pm$ 1.8 & 40.3 $\pm$ 2.2 & 21.4 $\pm$ 2.4 & 72.0 $\pm$ 2.6 & ~~~74 $\pm$ 1.7 & 43.8 $\pm$ 1.9 \\  \midrule
     \srawnsembles{MA+VP} & \textbf{66.9} $\pm$ 2.1 & ~~~52 $\pm$ 2.2 & \textbf{56.6} $\pm$ 2.2 & 38.6 $\pm$ 2.1 & \textbf{36.9} $\pm$ 2.8 & 68.9 $\pm$ 2.7 & 86.7 $\pm$ 1.3 & 40.7 $\pm$ 1.9 \\
     \srawnsembles{MA+TC} & \textbf{~~~67} $\pm$ 2.1 & 61.9 $\pm$ 2.1 & \textbf{56.4} $\pm$ 2.2 & 46.2 $\pm$ 2.2 & \textbf{37.2} $\pm$ 2.8 & 68.0 $\pm$ 2.7 & 87.2 $\pm$ 1.3 & 41.7 $\pm$ 1.9 \\
     \srawnsembles{VP+TC} & 53.1 $\pm$ 2.2 & ~~~63 $\pm$ 2.1 & 24.3 $\pm$ 1.9 & 46.6 $\pm$ 2.2 & 20.9 $\pm$ 2.3 & 67.4 $\pm$ 2.7 & 72.4 $\pm$ 1.7 & \textbf{44.3} $\pm$ 1.9 \\ \midrule
     \srawnsembles{All} & \textbf{66.6} $\pm$ 2.1 & \textbf{64.9} $\pm$ 2.1 & \textbf{55.7} $\pm$ 2.2 & \textbf{48.7} $\pm$ 2.2 & ~~~\textbf{36} $\pm$ 2.8 & 71.2 $\pm$ 2.6 & 86.2 $\pm$ 1.3 & 41.8 $\pm$ 1.9 \\
     \scatsembles{All} & 57.7 $\pm$ 2.2 & \textbf{66.9} $\pm$ 2.1 & 31.6 $\pm$ 2.1 & \textbf{49.9} $\pm$ 2.2 & 28.5 $\pm$ 2.6 & \textbf{75.4} $\pm$ 2.5 & 84.9 $\pm$ 1.4 & \textbf{44.6} $\pm$ 1.9 \\
     \bottomrule
\end{tabular}
\caption{Input- and output-based evaluation results of the methods and their ensembles, over different feature types and models, averaged across model layers, along with their respective 95\% confidence intervals. For SAE features we take the average over features from SAEs of all sizes. We denote \texttt{MA} for \maxact{}, \texttt{VP} for \vocabproj{}, \texttt{TC} for \tokenshift{}, and \texttt{EnsembleR} and \texttt{EnsembleC} for the raw and concatenation based ensembles.}
\label{tab:mean_results}
\end{table*}

\begin{figure*}[t]
  \centering
  \begin{subfigure}[b]{1\textwidth}
  \centering
  \includegraphics[width=0.9\textwidth]{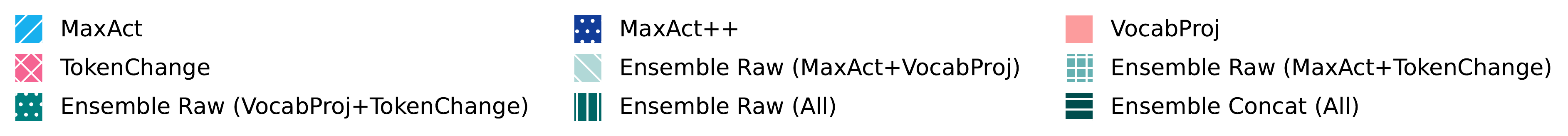}
  \label{fig:legend_models}
  \hspace*{-1em}
  \end{subfigure}
  \begin{subfigure}[b]{.49\textwidth}
    \includegraphics[width=\textwidth]{figures/Gemma_65k_res.pdf}
    \caption{Residual stream SAE features of width 65k from \gemma{}.}
  \end{subfigure}
  \hfill
  \begin{subfigure}[b]{.49\textwidth}
    \includegraphics[width=\textwidth]{figures/Gemma_65k_mlp.pdf}
    \caption{MLP SAE features of width 65k from \gemma{}.}
  \end{subfigure}

  \begin{subfigure}[b]{.49\textwidth}
    \includegraphics[width=\textwidth]{figures/Llama_32k_res.pdf}
    \caption{Residual stream SAE features of width 32k from \llama{}.}
  \end{subfigure}
  \hfill
  \begin{subfigure}[b]{.49\textwidth}
    \includegraphics[width=\textwidth]{figures/Transluce.pdf}
    \caption{MLP features from \llamaIF{}.}
  \end{subfigure}
  \caption{Performance of the various methods on the proposed metrics, for \gemmaF{} (upper row), \llamaF{} (lower left), and \llamaIF{} (lower right). For the output metric, the baseline (dashed black line) is $1/3$ since the judge LLM picks between three sets of texts.}
  \label{fig:result_plots}
\end{figure*}

\subsection{Results}
\label{sec:res}

Table ~\ref{tab:mean_results} shows the results averaged across layers, and Figure~\ref{fig:result_plots} provides a breakdown for layer groups for features from \gemma{} and both \llama{} models. Similar trends are shown for all other features in \S\ref{apx:results_gpt}.

\paragraph{Combining input- and output-centric methods yields better feature descriptions} Table~\ref{tab:mean_results} shows that across all models and feature types,
\maxact{} outperforms \vocabproj{} and \tokenshift{} on the input-based evaluation and vice versa on the output-based evaluation, often by large margins of up to 15\%-30\%. This also holds for \maxact{}++ on \llamaIF{}, demonstrating that input- and output-centric methods capture different feature information.
Second, ensembling input- and output-centric methods boosts performance on both evaluations, with the ensembles combining all three methods consistently outperforming the single-methods. For instance, for \gemma{} the ensembles yielded an improvement of 6\%-10\% over the next best single-method on both metrics. 
One exception to this trend is \maxact{}++, which performs better than all other methods on the input metric, with \rawnsemble{} in close second. This is probably due to \maxact{}++ being optimized for describing what activates a given feature. Overall, this input-output integration not only better describes the causal roles of features but also improves performance on the widely-used input-based evaluation.

\paragraph{Performance varies by layer and feature type}
Comparing the results for residual versus MLP features and neurons versus SAE features, we find that output-based performance is substantially lower for MLP features compared to residual features (reaching 45-50 points for MLP vs. $\sim$66 points for residual). 
This might be explained by the MLP layers introducing gradual changes to the residual stream \cite{geva-etal-2021-transformer, geva-etal-2022-transformer}, potentially making them harder to steer. Additionally, output-based performance of \vocabproj{} is worse in early layers but gradually improves, consistent with prior observations \cite{nostalgebraist2020interpreting, geva-etal-2021-transformer, yom-din-etal-2024-jump}.

\paragraph{\vocabproj{} and \tokenshift{} often provide efficient substitutes for \maxact{}}
A major practical drawback of \maxact{} is the computational cost required for comprehensively mapping the activating inputs of a feature. Considering the performance of \vocabproj{}, \tokenshift{}, and \srawnsembles{VP+TC}, we observe that (a)
they typically outperform \maxact{} on the output-based evaluation, which is crucial for assessing the description's faithfulness to the feature's causal effect and its usefulness for steering, and (b) they often perform only slightly worse on the input-based evaluation, e.g. there's only a 3.5 point gap between \rawnsembles{VP+TC} and \maxact{} on residual stream SAE features in \gemma. These results suggest that \vocabproj{} and \tokenshift{}, which require only $\leq$2 inference passes, can often be a more efficient and sometimes higher-performing alternative to the widely-used \maxact{} method. An analysis of the computational costs is in \S\ref{apx:compute_analysis}.

\paragraph{Description Format Affects Performance}
Comparing the top-performing ensembles, we observe that \rawnsemble{} is generally better on the input-based evaluation while \catsemble{} is consistently best on the output-side evaluation. 
We hypothesize that this could be due to the different description formats of the two ensembling approaches, i.e., concatenating raw outputs versus generated descriptions.
For the input-based evaluation, a longer and more informative description may have a higher chance of enabling an LLM to generate sentences with at least one activating token, compared to a concise description. Similarly, a concise description could be matched to texts generated by the model more easily compared to a long and detailed description.

\section{Analysis}
\label{sec:analysis}

In this section, we compare the feature descriptions obtained by \maxact{}, \vocabproj{} and \tokenshift{} and analyze the utility in their combination.  

\begin{table*}[t!]
    \setlength{\belowcaptionskip}{-10pt}
    \setlength{\tabcolsep}{3.5pt}
    \footnotesize
    \centering
    \begin{tabular}{p{1.5cm}p{2.2cm}p{3.8cm}p{7.5cm}}
        \toprule
        \textbf{Relation} & \textbf{Example feature} & \textbf{Description by \maxact{}} & \textbf{Description by \vocabproj{}} \\
         & layer-type/id &  & \\ 
        \toprule
        Similar \tcbox{41\%} & 
        \texttt{3-MLP-16K/ 4878}
        & Terms and themes related to various genres of storytelling, particularly in horror, drama, and fantasy. &
        A blend of themes and genres commonly found in storytelling or media, with a specific focus on dramatic, horror, and suspenseful narratives.
        \\ \midrule
        
        Composition \tcbox{23\%} & 
        \texttt{19-MLP-16K/ 5635}
        & References to political events and milestones. & 
        Concepts related to time measurement such as days, weeks, weekends, and months, indicating it likely pertains to scheduling or planning events.
        \\ \midrule

        Abstraction \tcbox{23\%} & 
        \texttt{21-RES-16K/ 10714}
        & Information related to bird species and wildlife activities. & 
        Concepts related to birdwatching and ornithology, focusing on activities such as observing, spotting, and recording bird species in their natural habitats. 
        \\ \midrule

        Different \tcbox{13\%} & 
        \texttt{19-MLP-16K/ 1450}
        & Mentions of notable locations, organizations, or events, particularly in various contexts. &
        Concepts related to self-reflection, purpose, and generalization in various contexts, 
        focusing on the exploration of identity and overarching themes in literature or philosophy.
        \\ \bottomrule
    \end{tabular}
    \caption{Human evaluation results of descriptions by \maxact{} and \vocabproj{} for 100 SAE features from Gemma Scope, showing for each relation category the fraction of observed cases and the descriptions of an example feature.}
    \label{table:descriptions_human_eval}
\end{table*}

\subsection{Qualitative Analysis}
\label{subsec:analysis}

We manually analyze the descriptions by \maxact{} and \vocabproj{} for a random sample of 100 features from \gemmasae{} 16K, 50 for the MLP layers and 50 for the residual stream. 
We exclude \tokenshift{} here as we noticed that the descriptions it produces are often similar to those by \vocabproj{} (see examples in \S\ref{apx:additional_quality}). In the analysis, we consider descriptions that pass both our input- and output-based evaluations. We observed 4 main types of relations between the descriptions:
\begin{itemize}
[itemsep=1pt, topsep=2pt,leftmargin=*]
\item \textbf{Similar}: The tokens in the projection and are highly similar to the tokens in the activating examples, resulting in matching descriptions.

\item \textbf{Composition}: The input- and output-centric descriptions refer to different aspects of the feature, while their composition provides a more holistic description of the feature. 

\item \textbf{Abstraction}: The tokens in the projection represent a more general or broad concept than the one observed in the activating examples.

\item \textbf{Different}: The input- and output-centric descriptions refer to different aspects of the feature, which share no clear relation between them. 
\end{itemize}
Table~\ref{table:descriptions_human_eval} shows the fraction of examples classified per category alongside representative feature descriptions. 
Overall, while input- and output-centric descriptions are often similar (41\%), there are many cases where their composition provides a broader (23\%) or more accurate (23\%) description .

\subsection{Reviving Dead Features}
\label{subsec:dead_neurons}

One drawback of describing features with \maxact{} is the dependency on the dataset used to obtain activations \cite{bolukbasi2021interpretabilityillusionbert}. A particularly interesting case is the classification of ``dead'' features, which do not activate for any input from the dataset.
Dead features can be prevalent \cite{voita-etal-2024-neurons, gao2024scalingevaluatingsparseautoencoders, templeton2024scaling}. For example, we observed they constitute up to 29\% of the features in some SAEs in \gemma{}.

While dead features could potentially not represent meaningful features, it may be that the dataset used simply does not cover the ``right'' inputs for activating them.
Here we conduct an analysis that shows that dead features can be ``revived'' (i.e. activated) with inputs crafted based on their \vocabproj{} and \tokenshift{} descriptions.

\paragraph{Analysis}
We sampled 1,850 SAE features from \gemmaF{} equally distributed across layers and types (MLP / residual) and classified as ``dead'' based on \pedia{}. For each feature, we create a set of candidate prompts for activating it by: (a) using the feature descriptions by \vocabproj{} and \tokenshift{} and letting Gemini generate 150 sentences that are likely to activate the feature, and (b) gathering the tokens identified by \vocabproj{} and \tokenshift{} and constructing 1,450 sequences of different lengths that randomly combine these tokens. Both the top and bottom tokens obtained using these methods could potentially activate the feature, as they might relate to concepts that the feature promotes or suppresses.
We then feed all the generated prompts into the model and consider a feature as ``revived'' if any prompt successfully activated it. For implementation details, see \S\ref{apx:dead_neurons}.

\paragraph{Results} The generated prompts successfully activated 9.1\% (85) of MLP SAE features and 62\% (491) of residual ones. In 12\% (70) of cases, a feature was activated using an LLM-generated prompt, while 73\% (423) were activated with a prompt composed of two tokens: `\texttt{<BOS>}' and a sampled token. Moreover, the revived dead features can often be easily interpreted using \vocabproj{} and \tokenshift{}, while considered faithful based on our output-based metric (see examples in \S\ref{apx:dead_neurons}).
Overall, this demonstrates that output-centric methods
can address potential oversights that may arise from focusing solely on activating inputs.

\section{Related Work}

\citet{bills2023language} introduced an automated interpretability pipeline that used 
GPT-4 to explain the neurons of GPT-2 based on their activating examples (\maxact{}), while employing an input-based evaluation known as simulation scoring. This approach has become common practice for interpreting neurons and learned SAE features of LLMs at scale \cite{neuronpedia, cunningham2023sparse, bricken2023monosemanticity, templeton2024scaling, gao2024scalingevaluatingsparseautoencoders, he2024llamascopeextractingmillions}, which also extends to neuron description pipelines of visual models \cite{hernandez2022natural, shaham2024a, kopf2024cosyevaluatingtextualexplanations}.

Recently, new methods for generating feature descriptions have been proposed, such as applying variants of activation patching \cite{kharlapenko-2024}, refining the prompt given to the explainer model \cite{paulo2024automaticallyinterpretingmillionsfeatures}, and improving descriptions of residual feature activations via description selection \cite{choi2024automatic} similarly to the algorithm by \citet{singh2023explaining}.
While all these prior works rely on input-centric, computationally intensive approaches, we propose output-centric efficient methods that require no more than two inference passes of the model. Furthermore, we show that combining input- and output-centric methods leads to improved overall performance.

More broadly, our work relates to growing efforts in understanding features encoded in neurons and SAE features.
These include steering \cite{farrell2024applying, chalnev2024improvingsteeringvectorstargeting, o2024steering, templeton2024scaling}, circuit discovery \cite{marks2024sparsefeaturecircuitsdiscovering, makelov2024towards, balcells2024evolutionsaefeatureslayers}, feature disentanglement \cite{huang-etal-2024-ravel, cohen2024evaluating} and benchmarks like SAEBench.\footnote{\url{https://www.neuronpedia.org/sae-bench/info}}
However, evaluation of feature descriptions remains relatively underexplored.
\citet{rajamanoharan2024jumpingaheadimprovingreconstruction} evaluated latent interpretability for different SAE architectures using an input-centric approach which does not reflect downstream effect in model control.
More recently, \citet{paulo2024automaticallyinterpretingmillionsfeatures} have found negative correlation between multiple input-centric scoring methods and an intervention-based metric. Finally, \citet{bhalla2024unifyinginterpretabilitycontrolevaluation} concurrently evaluated feature descriptions in terms of their downstream effects on the model. However, they focus on evaluating methods for effectively steering models, as opposed to evaluating methods for generating descriptions.

\section{Conclusion}

While existing automated interpretability efforts describe features based on their activating inputs, 
we posit that describing a feature is a two-faceted challenge, requiring the comprehension of both its activating inputs and influence on model outputs. To tackle this challenge at scale, we employ two evaluations -- input-based and output-based -- and propose two output-centric methods (\vocabproj{} and \tokenshift{}) for generating feature descriptions. Through extensive experiments we show that output-centric methods offer an efficient solution for automated interpretability, especially when geared towards model steering, and can substantially enhance existing pipelines which rely on input-centric methods.

\section*{Limitations}

Although we observe clear trends in the results, the output-based evaluation is fairly noisy. We address this by sampling large numbers of features and using multiple prompts in the evaluation, but future work could focus on reducing this noise further and making the evaluation more efficient. 
Additionally, we find that the output-centric methods and ensembles are sensitive to the choice of prompt. Since generating feature descriptions using these methods is non-trivial and often involves long texts (especially for the ensembles), improving explainer model prompts to extract relevant information could potentially enhance performance. We also note that our input-based evaluation uses a binary threshold, which may oversimplify feature behavior. Nonetheless, it enabled us to efficiently identify trends across models and methods, and we leave refining this evaluation to future work.

Regarding the methods evaluated, while we focused on efficient approaches that can automatically scale to millions of features, exploring other methods, such as patching-based methods, could be valuable. 
Lastly, the output-centric methods we propose are tied to the model's vocabulary, which means they can only describe features that can be expressed with tokens from the vocabulary. These methods may struggle in describing features that are not easily or naturally expressed with words, such as positional features. For simplicity, we did not differentiate between whether concepts were being suppressed or promoted by a feature.

\section*{Acknowledgements}
We thank the Transluce team, specifically Dami Choi, for sharing their neuron description pipeline data, as well as Johnny Lin from \pedia{} for sharing their descriptions and model activations data.
This work was supported in part by the Gemma 2 Academic Research Program at Google, the Edmond J. Safra Center for Bioinformatics at Tel Aviv University, a grant from Open Philanthropy, and the Israel Science Foundation grant 1083/24.
Figures~\ref{fig:intro} and~\ref{fig:evals} use images from \url{www.freepik.com}.

\bibliography{custom}

\appendix

\section{Additional Details on Feature Description Evaluations}
\label{apx:evals}

\begin{table*}[t]
    \setlength{\tabcolsep}{3.5pt}
    \footnotesize
    \centering
    \begin{tabular}{llll}
        \toprule
        Variant & \gemmaF{}         & \llamaF{}         & \gpt{}          \\
                & (\gemmasae{} 16K) & (\llamasae{} 32K) & (\gptsae{} 32K) \\ 
        \toprule
        Dec \& Unembed & 0.44 (0.31-0.58) & 0.27 (0.17-0.39) & 0.29 (0.12-0.50)
        \\ 
        Enc \& Unembed & 0.38 (0.27-0.52) & 0.14 (0.06-0.25) & 0.25 (0.12-0.46)
        \\ 
        Dec \& Embed & 0.52 (0.38-0.65) & 0.20 (0.11-0.31) & 0.25 (0.08-0.46)
        \\ 
        Enc \& Embed & 0.29 (0.17-0.42) & 0.16 (0.08-0.27) & 0.21 (0.08-0.42)
        \\ \bottomrule
    \end{tabular}
    \caption{Confidence interval of mean input metric results on the descriptions generated by \vocabproj{} using tokens retrieved by 4 different methods, to compare decoding vs. encoding variants.}
    \label{table:dec_enc_results}
\end{table*}

\begin{table}[t]
    \footnotesize
    \centering
    \begin{tabular}{lp{1.3cm}p{1.3cm}p{1.3cm}}
        \toprule
        Method & \multicolumn{3}{c}{Estimated FLOPs} \\
        & \gemmaF{} & \gemmaNF{} & \gemmaTF{} \\
        \toprule
        \maxact{}    & $3.9 \cdot 10^{16}$ & $1.5 \cdot 10^{17}$ & ~~~$5 \cdot 10^{17}$ \\
        \vocabproj{}  & $2.8 \cdot 10^{14}$ & $1.1 \cdot 10^{15}$ & ~~~$2 \cdot 10^{15}$ \\ 
        \tokenshift{} & $9.9 \cdot 10^{13}$ & $4.1 \cdot 10^{14}$ & $1.3 \cdot 10^{15}$ \\
        \bottomrule
    \end{tabular}
    \caption{Estimated FLOPs for generating descriptions for all MLP features for models of different sizes, on a sample of 25k sequences of 128 tokens each, as done by \pedia{}.}
    \label{table:compute_analysis}
\end{table}

\begin{table*}[t]
    \footnotesize
    \centering
    \begin{tabular}{llll}
        \toprule
        Variant & \gemmaF{}         & \llamaF{}         & \gpt{}          \\
                & (\gemmasae{} 16K) & (\llamasae{} 32K) & (\gptsae{} 32K) \\ 
        \toprule
        Dec \& Unembed & 0.41 (0.35-0.47) & 0.14 (0.11-0.19) & 0.20 (0.13-0.28)
        \\ 
        Dec \& Embed & 0.37 (0.31-0.43) & 0.12 (0.08-0.16) & 0.13 (0.08-0.21)
        \\ \bottomrule
    \end{tabular}
    \caption{Confidence interval of mean input metric results on the descriptions generated by \vocabproj{} using tokens retrieved by 2 different methods, to compare unembedding vs. embedding variants with the decoding matrix.}
    \label{table:unembed_embed_results}
\end{table*}

\begin{figure}[t]
    \centering
    \includegraphics[scale=0.85]{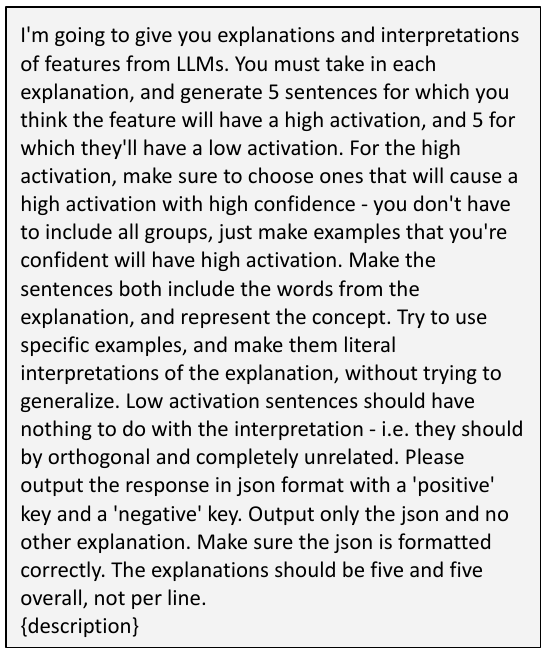}
    \caption{Prompt given to the judge LLM for the input-based evaluation.}
    \label{fig:prompt1}
\end{figure}

\paragraph{Input-based}
We used the prompt in Figure~\ref{fig:prompt1} for generating activating and neutral sentences based on a feature description, as per the input metric.

\begin{figure}[t]
    \centering
    \includegraphics[scale=0.85]{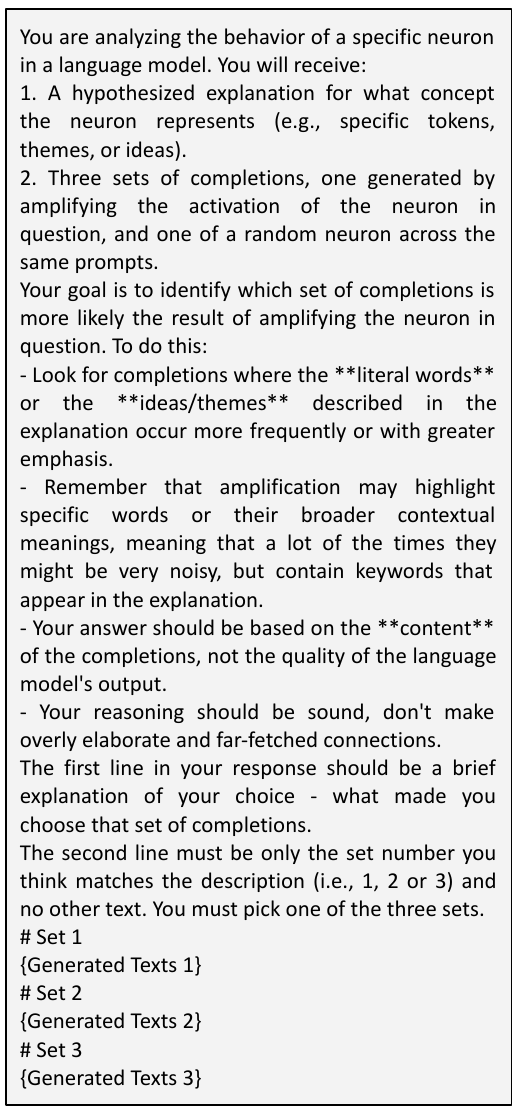}
    \caption{Prompt given to the judge LLM for the output-based evaluation.}
    \label{fig:prompt2}
\end{figure}

\begin{figure}[t]
    \centering
    \includegraphics[scale=0.85]{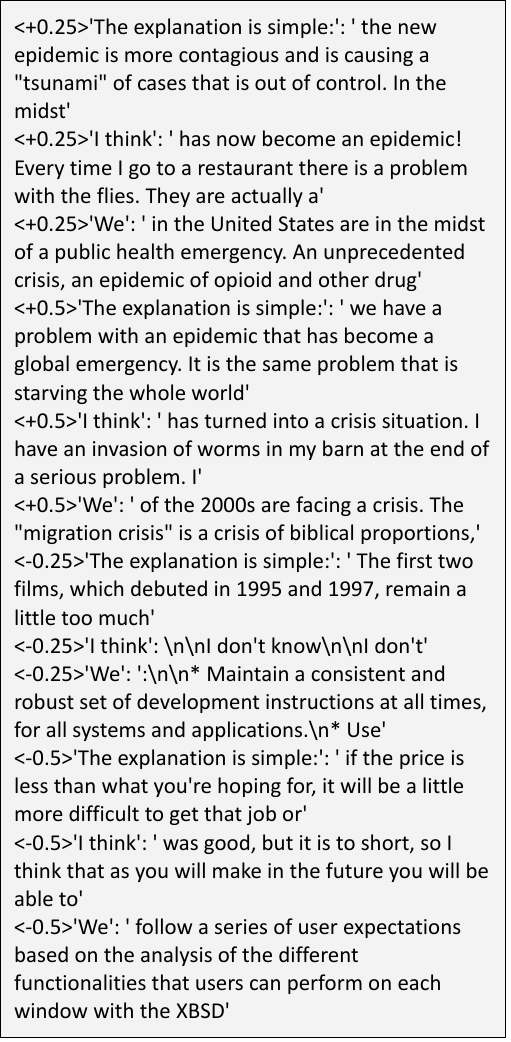}
    \caption{An example of a steered text set for the output-based metric.}
    \label{fig:prompt2_user}
\end{figure}

\paragraph{Output-based}
We used the prompt in Figure~\ref{fig:prompt2} for tasking the judge LLM with telling the steered text generations apart using a feature description, as per the output metric. Figure~\ref{fig:prompt2_user} shows an example of a steered text set for the feature with the description \texttt{``urgent global issues such as epidemics and invasions''}.

\begin{figure}[t]
    \centering
    \includegraphics[scale=0.85]{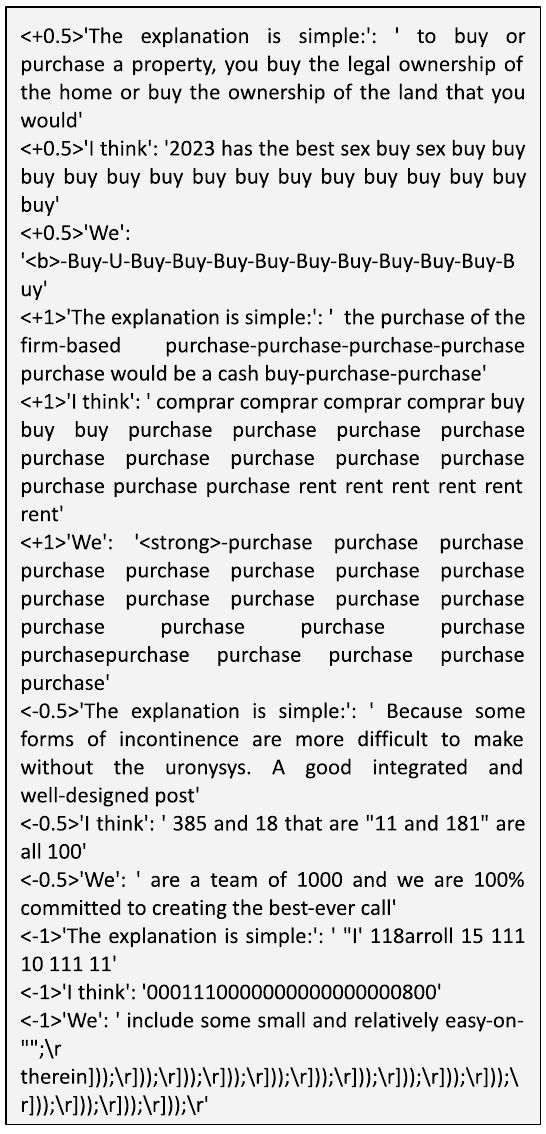}
    \caption{Higher clamping value when steering feature with description ``purchasing activities, including buying, viewing, and downloading products'', leading to degenerate text.}
    \label{fig:degenerate}
\end{figure}

The clamping values for $m$ were derived by fixing two target KL-divergence values, $0.25$ and $0.5$, providing two positive and two negative clamping values for $m$. These values, along with sequence length, balance generating text with sufficient feature effect, and producing long or degenerate text that is difficult to evaluate.
To confirm this, we ran the output-based evaluation using target KL-divergence values of $0.5$ and $1$ on 800 features from \gemmaF{}, obtaining similar results. However, the generated text became more degenerate (see Figure~\ref{fig:degenerate} for an example text). Therefore, we decided to retain the original target KL-divergence values, as higher values resulted in text that did not reflect probable model behavior.

\section{Additional Experimental Details}
\label{apx:experimental}

\subsection{Variants of \vocabproj{}}
\label{subsec:vocabproj_variants}
When implementing \vocabproj{}, presented in \S\ref{sec:generation}, there are several variants that generate tokens we can choose from, which are determined by the weight matrices we utilize. There are two points of interest: (a) the projection destination in the model (\textit{unembedding} matrix $W_U \in \mathbb{R}^{d \times |\mathcal{V}|}$ vs. \textit{embedding} matrix $W_E \in \mathbb{R}^{|\mathcal{V}| \times d}$), (b) in the case of SAEs, the source of the feature vector we analyze when applying the SAE on the hidden representation (\textit{encoding} matrix $W_{enc} \in \mathbb{R}^{d \times d_{sae}}$ vs. \textit{decoding} matrix $W_{dec} \in \mathbb{R}^{d_{sae} \times d}$). We conducted experiments across all of our subject models (except \llamaIF{}), in order to choose the best variant of this method.

\paragraph{Decode vs. Encode} We first wished to tackle the decision of (a). To do so, we conducted a small-scale experiment in which we took a random sample of SAE features, using the following SAE types: \gemmasae{} 16K, \llamasae{} 32K and \gptsae{} 32K; considering both layers (MLP and residual), for each subject model. This resulted in 52 features from \gemmaF{}, 64 from \llamaF{}, and 24 from \gpt{}. Due to the small sample size of features, we used bootstrap (9999 resamples of the data with replacements, 95\% confidence) to estimate the accuracy of each variant.
We used our chosen prompt (see \S\ref{subsec:description_generation} for more details), to generate descriptions given the tokens retrieved using each of the 4 combinations above. We evaluated the descriptions using our input metric presented in \S\ref{sec:scoring}. 
Table~\ref{table:dec_enc_results} shows the confidence interval for each variant on each model. From the table we concluded that generally the decoding matrix variant outperforms the encoding one.

\paragraph{Unembed vs. Embed} We then conducted a larger scale experiment to tackle decision (b). We used the same SAEs and models from our previous experiment, taking a random sample of 5 features per SAE, considering both layers (MLP and residual), for each subject model. This resulted in 260 features from \gemmaF{}, 320 from \llamaF{}, and 120 from \gpt{}.
Table~\ref{table:unembed_embed_results} shows the confidence interval for each variant on each model. From the table we concluded that the unembedding variant outperforms the embedding one, therefore we chose the decoding-unembedding variant for \vocabproj{}.

\begin{table*}[t]
    \footnotesize
    \centering
    \begin{tabular}{lllll}
        \toprule
        Variant  & \gemmaF{}         & \llamaF{}         & \gpt{}          & ALL \\
                 & (\gemmasae{} 16K) & (\llamasae{} 32K) & (\gptsae{} 32K) & \\ 
        \toprule
        \pedia{} & 0.49 (0.44-0.55) & 0.46 (0.41-0.52) & 0.41 (0.36-0.47) & 0.46 (0.43-0.49)
        \\ 
        \maxact{} & 0.52 (0.46-0.57) & 0.47 (0.42-0.53) & 0.44 (0.39-0.50) & 0.48 (0.45-0.51)
        \\ \bottomrule
    \end{tabular}
    \caption{Confidence interval of mean input metric results on the descriptions taken from \pedia{} and those generated by \maxact{}.}
    \label{table:input_recreate}
\end{table*}

\begin{figure}[t]
  \centering
  \begin{subfigure}[b]{1\textwidth}
  \includegraphics[width=0.49\textwidth]{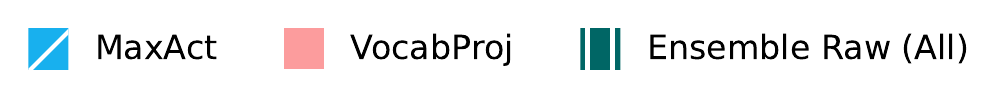}
  \label{fig:small_legend}
  \end{subfigure}
  \begin{subfigure}[b]{.49\textwidth}
    \includegraphics[width=\textwidth]{figures/human_eval.pdf}
  \end{subfigure}
  \vspace*{-2.6em}
  \caption{Human evaluation results for 100 features across three methods, for input- and output-faithfulness.}
  \label{fig:human_eval}
\end{figure}

\begin{figure}[t]
    \centering
    \includegraphics[width=0.99\columnwidth]{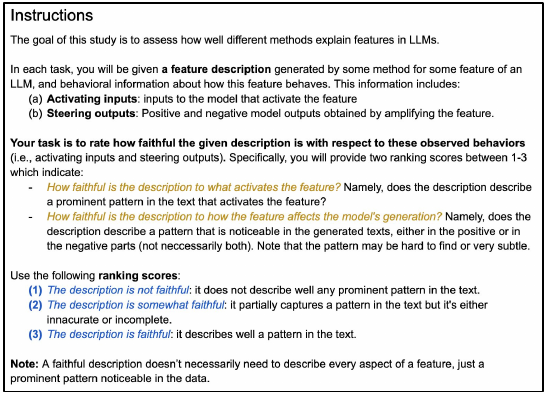}
    \caption{Instructions provided to human annotators for the evaluation of feature descriptions. These were accompanied with a few example annotations.}
    \label{fig:instructions}
\end{figure}

\begin{figure}[t]
    \centering
    \includegraphics[scale=0.85]{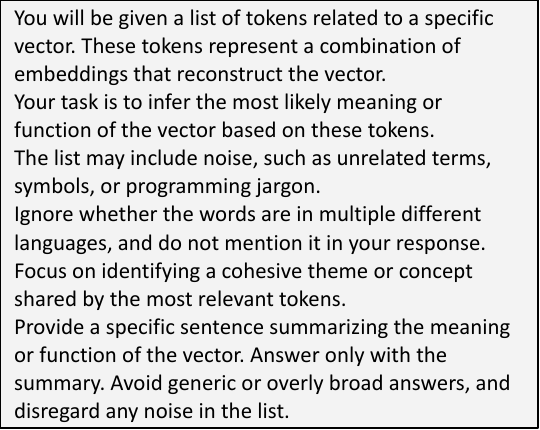}
    \caption{Prompt given to the explainer model for the \vocabproj{} method.}
    \label{fig:prompt3}
\end{figure}

\subsection{Description Generation}
\label{subsec:description_generation}

\paragraph{\vocabproj{}} 
We use the prompt in Figure~\ref{fig:prompt3} given to the explainer model for it to generate feature descriptions using \vocabproj{}.

\begin{figure*}[t]
    \centering
    \includegraphics[scale=0.6]{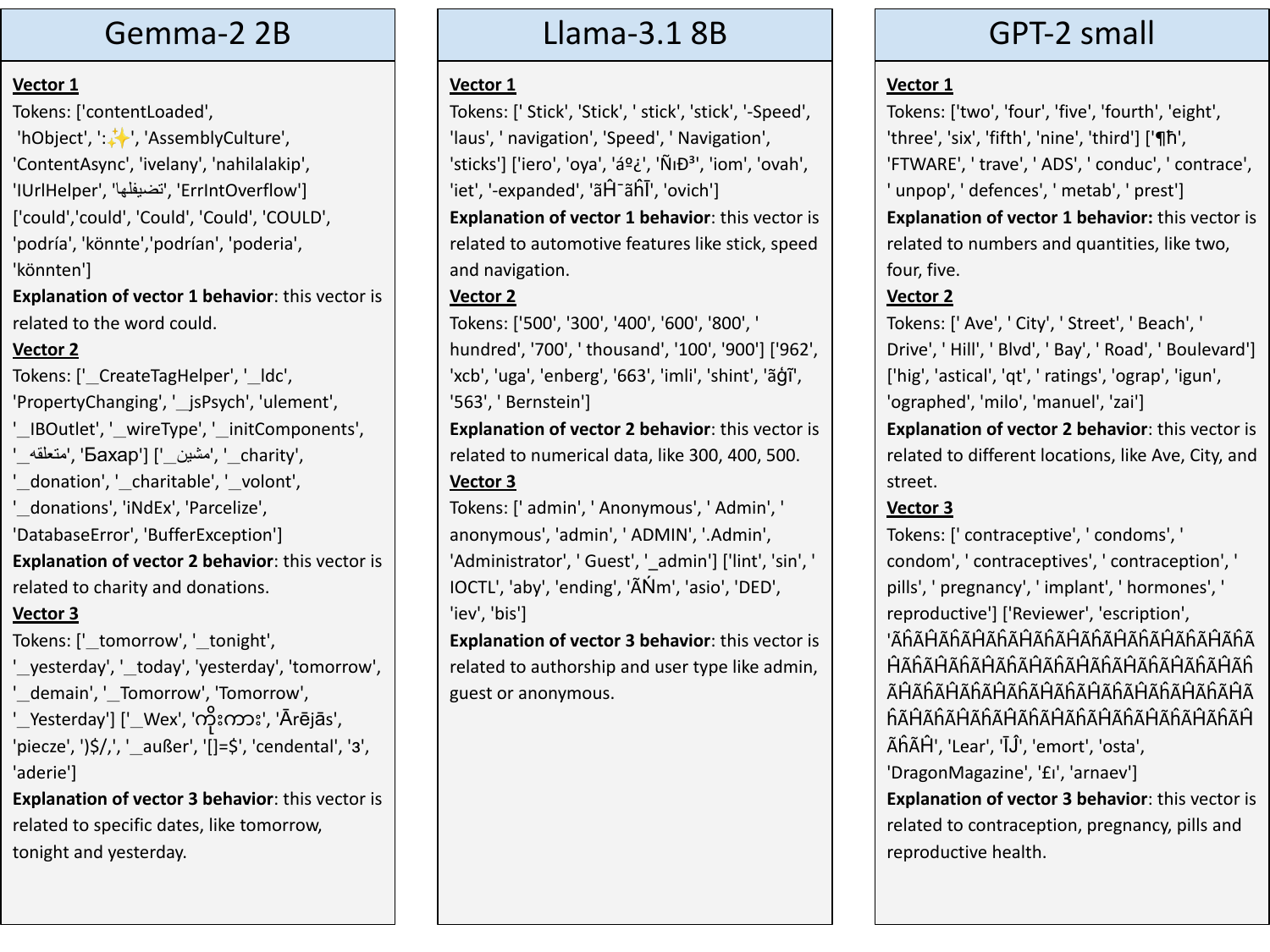}
    \caption{Three demonstrations of tokens and their descriptions for each model, added to the base prompt forming a fine-tuned prompt.}
    \label{fig:ftp}
\end{figure*}

\begin{figure}[t]
    \centering
    \includegraphics[scale=0.85]{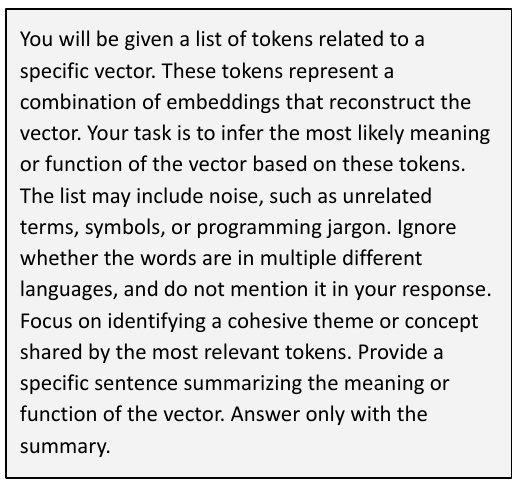}
    \caption{The basic fine-tuned prompt \vocabproj{} method.}
    \label{fig:prompt6}
\end{figure}

\begin{figure}[t]
    \centering
    \includegraphics[scale=0.85]{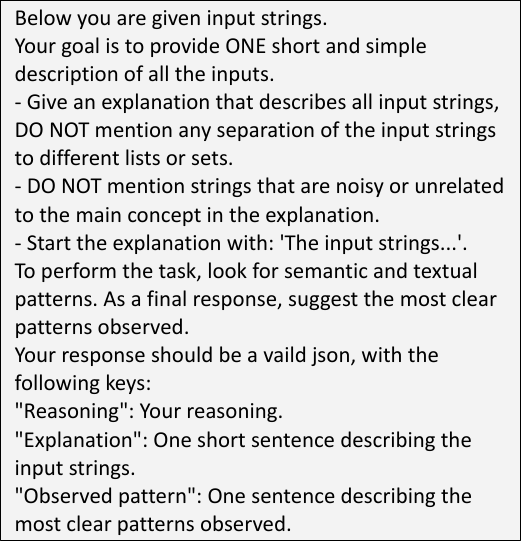}
    \caption{A first variant of a generic prompt for the \vocabproj{} method.}
    \label{fig:prompt4}
\end{figure}

\begin{figure}[ht]
    \centering
    \includegraphics[scale=0.85]{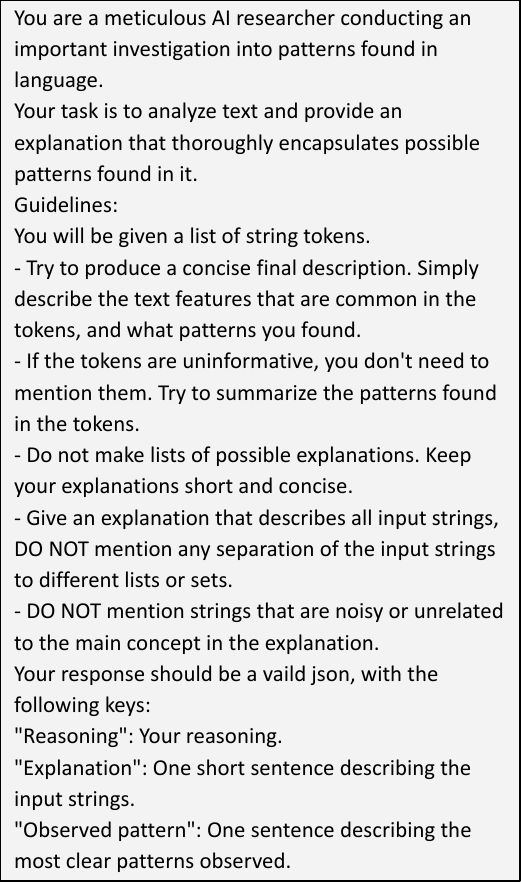}
    \caption{A second variant of a generic prompt for the \vocabproj{} method.}
    \label{fig:prompt5}
\end{figure}

We tried different prompts, but didn't observe significant improvement. These include
both generic prompts to be used for all subject models (Figures~\ref{fig:prompt4} and~\ref{fig:prompt5}), and more fine-tuned prompts based on vocabulary projection demonstrations for each subject model (see the fine-tuned based prompt in Figure~\ref{fig:prompt6}, for which we concatenate few-shot examples for each model as seen in Figure~\ref{fig:ftp}).

\begin{figure*}[t]
    \centering
    \includegraphics[scale=0.7]{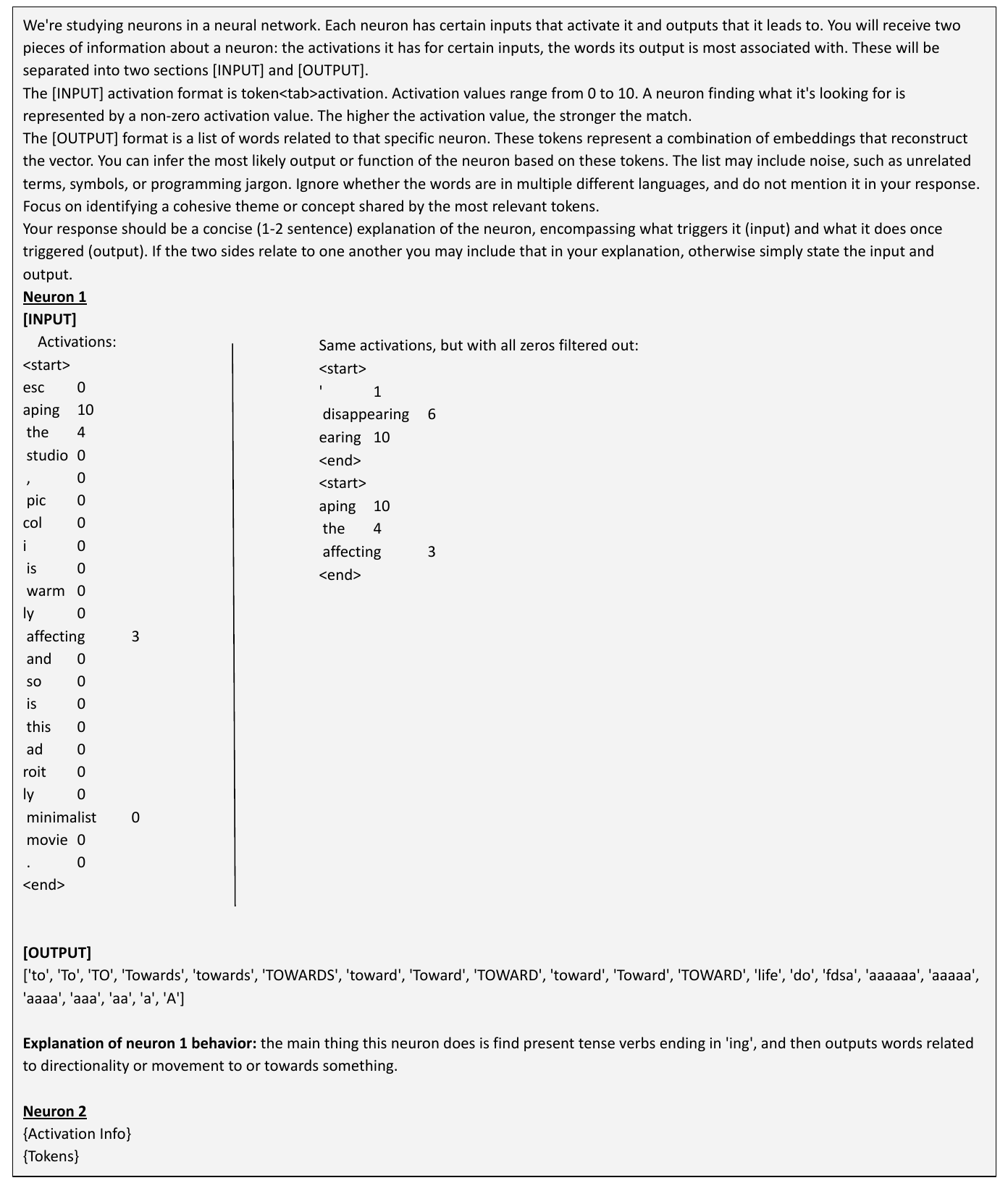}
    \caption{Prompt given to the explainer model for the \rawnsemble{} method.}
    \label{fig:prompt7}
\end{figure*}

\paragraph{Ensembles}
To generate \rawnsemble{} descriptions, we used variations of the prompt in Figure~\ref{fig:prompt7} when the ensemble included \maxact{}. To generate \rawnsembles{\vocabproj{}+\tokenshift{}} we simply concatenate the tokens generated by the two methods and use the \vocabproj{} prompt.

\subsection{Recreating \pedia{} Descriptions using \maxact{}}
\label{subsec:recreating_pedia}
In order to compare our own generated descriptions to the ones provided in \pedia{}, we conducted an experiment across all of our subject models (except \llamaIF{}) where we regenerated a description based on the activations data provided by \pedia{}, fed to \maxact{}, 
following their automatic pipeline based on \citet{bills2023language}. For a given feature, the explainer model gets as input the 5 top-activating sentences in the format of token-activation pairs, and generates a description adapting their code\footnote{\url{https://github.com/hijohnnylin/automated-interpretability}} to our pipeline.

We took a random sample of 360 SAE features from each model, using the following SAE types: \gemmasae{} 16K and 65K, \llamasae{} 32K and \gptsae{} 32K and 128K; considering both layers (MLP and residual). We evaluated both sets of descriptions using our input-based metric, and observed that they reach similar performance. Table~\ref{table:input_recreate} shows the confidence interval for the mean input metric evaluating both \pedia{}'s descriptions and our recreated descriptions. 

\begin{table*}[t]
\centering
\footnotesize
\begin{tabular}{lcccccccc}
\toprule
      & \multicolumn{2}{c}{\textbf{\llama{} MLP SAE}} & \multicolumn{2}{c}{\textbf{GPT2 Res. SAE}} & \multicolumn{2}{c}{\textbf{GPT2 MLP SAE}} \\
     & Input & Output & Input & Output & Input & Output \\ \midrule
     \maxact{} & 56.4 $\pm$ 2.9 & 49.6 $\pm$ 2.9 & 44.4 $\pm$ 2.3 & 44.1 $\pm$ 2.3 & 39.7 $\pm$ 2.9 & ~~~34 $\pm$ 2.8 \\
     \vocabproj{} & 20.2 $\pm$ 2.3 & 48.2 $\pm$ 2.9 & 23.7 $\pm$ 2~~~ & 42.8 $\pm$ 2.3 & ~~6.3 $\pm$ 1.4 & \textbf{38.3} $\pm$ 2.9 \\
     \tokenshift{} & 25.4 $\pm$ 2.5 & \textbf{53.1} $\pm$ 2.9 & 25.4 $\pm$ 2.1 & 43.4 $\pm$ 2.3 & ~~6.1 $\pm$ 1.4 & 36.5 $\pm$ 2.8 \\  \midrule
     \srawnsembles{MA+VP} & 62.1 $\pm$ 2.8 & 45.8 $\pm$ 2.9 & \textbf{59.6} $\pm$ 2.3 & \textbf{47.2} $\pm$ 2.4 & \textbf{51.2} $\pm$ 2.9 & \textbf{38.1} $\pm$ 2.9 \\
     \srawnsembles{MA+TC} & \textbf{65.8} $\pm$ 2.8 & 48.9 $\pm$ 2.9 & \textbf{58.8} $\pm$ 2.3 & \textbf{47.2} $\pm$ 2.4 & \textbf{51.1} $\pm$ 2.9 & \textbf{40.3} $\pm$ 2.9 \\
     \srawnsembles{VP+TC} & 22.6 $\pm$ 2.4 & 50.7 $\pm$ 2.9 & 29.2 $\pm$ 2.1 & 44.2 $\pm$ 2.4 & ~~7.1 $\pm$ 1.5 & \textbf{40.9} $\pm$ 2.9 \\ \midrule
     \srawnsembles{All} & 62.7 $\pm$ 2.8 & 51.6 $\pm$ 2.9 & \textbf{60.4} $\pm$ 2.3 & \textbf{47.2} $\pm$ 2.4 & \textbf{50.2} $\pm$ 2.9 & 37.1 $\pm$ 2.8 \\
     \scatsembles{All} & 39.1 $\pm$ 2.8 & \textbf{55.5} $\pm$ 2.9 & 42.4 $\pm$ 2.3 & \textbf{46.9} $\pm$ 2.4 & 24.4 $\pm$ 2.5 & 37.2 $\pm$ 2.8 \\
     \bottomrule
\end{tabular}
\caption{Input- and output-based evaluation results of the methods and their ensembles, over different feature types and models, averaged across model layers, along with their respective 95\% confidence intervals. For \gpt{} SAE features we take ones with width 32k. We denote \texttt{MA} for \maxact{}, \texttt{VP} for \vocabproj{}, \texttt{TC} for \tokenshift{}, and \texttt{EnsembleR} and \texttt{EnsembleC} for the raw and concatenation based ensembles.}
\label{tab:extra_mean_results}
\end{table*}

\begin{figure}[t]
    \centering
    \includegraphics[scale=0.85]{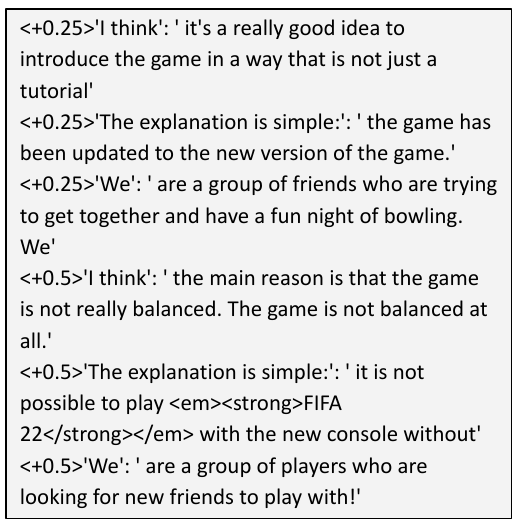}
    \caption{Text generated when amplifying a feature pronounced to be dead, which we managed to activate using the explanation generated by \vocabproj{}, which was ``gaming, focusing on players, gameplay, and game mechanics''.}
    \label{fig:dead_example}
\end{figure}

\section{Additional Evaluation Results}
\label{apx:results_gpt}
\begin{figure*}[t]
  \centering
  \begin{subfigure}[b]{1\textwidth}
  \centering
  \includegraphics[width=0.9\textwidth]{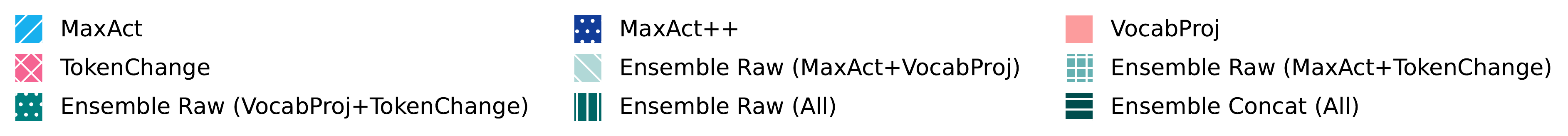}
  \label{fig:extra_legend_models}
  \hspace*{-1em}
  \end{subfigure}
  \begin{subfigure}[b]{.49\textwidth}
    \includegraphics[width=\textwidth]{figures/Llama_32k_mlp.pdf}
    \caption{MLP 32k SAE features from \llama{}.}
  \end{subfigure}
  \hfill
  \begin{subfigure}[b]{.49\textwidth}
    \includegraphics[width=\textwidth]{figures/gpt_32k_resid-mid.pdf}
    \caption{Mid residual stream 32k SAE features from \gpt{}.}
  \end{subfigure}

  \begin{subfigure}[b]{.49\textwidth}
    \includegraphics[width=\textwidth]{figures/gpt_32k_resid-post.pdf}
    \caption{Residual stream 32k SAE features from \gpt{}.}
  \end{subfigure}
  \hfill
  \begin{subfigure}[b]{.49\textwidth}
    \includegraphics[width=\textwidth]{figures/gpt_32k_mlp-out.pdf}
    \caption{MLP 32k SAE features from \gpt{}.}
  \end{subfigure}

  \caption{Performance of the various methods on the proposed metrics, for \llamaF{} (upper left) and \gpt{} (upper right and lower row). For the output metric, the baseline (dashed black line) is $1/3$ since the judge LLM picks between three sets of texts.}
  \label{fig:extra_result_plots}
\end{figure*}

See Figure~\ref{fig:extra_result_plots} and Table~\ref{tab:extra_mean_results} for additional results from \llamaF{} and \gpt{} SAE features, overall following the same trends observed in \S\ref{sec:exp}. Results for \gpt{} are noisier than in other models. This may be due to the model's relatively small size and generally lower performance.

\section{Computational Cost Analysis}
\label{apx:compute_analysis}
The computational cost of each method is a key factor to consider when selecting a method for generating descriptions. In our analysis, we computed the FLOPs required by each method to generate a description for every single MLP feature in \gemmaF{} (results in Table~\ref{table:compute_analysis}). When calculating the FLOPs required for a single forward pass, we rely on the heuristic FLOPs $\approx$ 6N plus embedding FLOPs, where N is the total number of non-embedding model parameters \cite{kaplan2020scalinglawsneurallanguage, anil2023palm2technicalreport}. The results show that even when using a small sample for \maxact{}---25k sequences of 128 tokens each, as used by \pedia{}---alternative methods are 2-3 orders of magnitude more compute-efficient. When using larger samples that more accurately represent a model's training data, such as The Pile \cite{gao2020pile800gbdatasetdiverse}, the difference reaches 7-8 orders of magnitude. Lastly, computational cost for analysing SAE features results in an increase of roughly one order of magnitutde across the board, while maintaining the same relative differences between methods.

\section{Human Evaluations}
\label{apx:human_eval}
To lend credence to our use of an LLM-judge and assess how well LLM-generated feature descriptions align with human judgment, we conducted two human evaluations.

\subsection{Justifying Using an LLM as a Judge}
\label{subsec:llm_judge_justification}
To justify our use of an LLM-as-a-judge in the output-based evaluation, we apply the alternative annotator test proposed by \citet{calderon2025alternativeannotatortestllmasajudge}. Following their procedure, we use three human annotators (graduate students) and a set of 100 randomly selected feature examples, evenly split between \llamaF{} and \gemmaF{}. For each feature, human annotators were given feature descriptions generated by \vocabproj{}, and the three text sets $\mathcal{T}_{\vv}$, $\mathcal{T}_{\vv'}$, and $\mathcal{T}_{\vv''}$. Each annotator then indicated which of the three sets matches the given description, as per the output-based metric. 
Consistent with \citet{calderon2025alternativeannotatortestllmasajudge}, we set $\epsilon=0.15$ to reflect our use of graduate student annotators. The analysis yielded a winning rate $\omega=1$ with p-value 0.03, supporting our use of an LLM-as-a-judge.

\subsection{Evaluating LLM Generated Descriptions}
\label{subsec:human_desc_eval}

To evaluate how well LLM-generated feature descriptions align with human judgment, we tasked human annotators (6 graduate students) with scoring their faithfulness with respect to (a) input-faithfulness: what activates the feature and (b) output-faithfulness: how the feature affects the model's outputs. The instructions provided to the annotators are shown in Figure~\ref{fig:instructions}.
We collected annotations for feature descriptions generated by \maxact{}, \vocabproj{}, and \rawnsembles{All} for 100 randomly selected SAE features from \gemmaF{}.

Figure~\ref{fig:human_eval} shows the results, where the overall trends align well with our proposed input- and output-based evaluations, discussed in \S\ref{sec:res}. \maxact{} performs better on the input evaluation, \vocabproj{} on the output evaluation, and \rawnsembles{All} performs best on both. However, \vocabproj{} performed slightly worse than expected on the output evaluation.  This discrepancy may stem from the difficulty humans face in evaluating a feature’s effect on text generation, as it requires detecting subtle changes across multiple texts. Indeed, in the annotator test conducted in \S\ref{apx:human_eval}, the judge LLM outperformed human annotators, supporting this claim. Furthermore, \maxact{}'s success in the input evaluation could be influenced by the descriptions being derived from the same data used for comparison, potentially biasing results in its favor. Nonetheless, these findings reinforce the claims in \S\ref{sec:res}, that input-centric methods perform better on input-based evaluations, output-centric methods on output-based ones, and ensembles perform best on both.

\section{Additional Details and Examples for Dead Feature Analysis}
\label{apx:dead_neurons}

\subsection{Generating Candidate Prompts}
To generate the  candidate prompts, we first generate 150 potentially activating sentences in the same way as when doing so for the output metric, based on \vocabproj{} and \maxact{}. We then compile a list of tokens using both \vocabproj{} and \tokenshift{}, and create candidate prompts that begin with \texttt{<BOS>} followed by either of the following:
\begin{itemize}
[itemsep=1pt, topsep=2pt,leftmargin=*]
    \item A single token (1 candidate per token).
    \item Two random tokens (250 candidates).
    \item Three random tokens (250 candidates).
    \item Five random tokens (200 candidates).
    \item Twelve random tokens (200 candidates).
    \item Twenty-five random tokens (100 candidates).
    \item Thirty-two random tokens (50 candidates).
\end{itemize}

\subsection{Dead Feature Revival Example}
As an example of a feature deemed to be dead that we managed to revive, and that also has a clear and faithful description, we take residual stream SAE feature 64628 in layer 23 of \gemmaF{}. Using \vocabproj{} we can get an explanation for the feature: \texttt{``gaming, focusing on players, gameplay, and game mechanics''}. Indeed when examining the top tokens when projecting the feature to vocabulary space, they are all related to games, and players. The candidate prompt that managed to trigger this feature is \texttt{``**Player Agency**: Choices and consequences, branching narratives.''}. We can then see in Figure~\ref{fig:dead_example} that this description is faithful when amplifying the feature and examining text generated from open ended prompts, like in the output evaluation.

\section{Additional Examples for Qualitative Analysis}
\label{apx:additional_quality}

Table~\ref{table:additional_human_eval} shows descriptions generated by \maxact{}, \vocabproj{} and \tokenshift{}.

\begin{table*}[t]
\setlength{\tabcolsep}{3.5pt}
\footnotesize
\centering
\resizebox{0.99\textwidth}{!}{
\begin{tabular}{p{2.2cm}p{3.5cm}p{4.8cm}p{4.8cm}}
    \toprule
     \textbf{Example feature} & \textbf{Description by \maxact{}} & \textbf{Description by \vocabproj{}} & \textbf{Description by \tokenshift{}{}}\\
     layer-type/id & & & \\ 
    \toprule
    
    \texttt{3-MLP-16K/ 4878}
    & Terms and themes related to various genres of storytelling, particularly in horror, drama, and fantasy. &
    A blend of themes and genres commonly found in storytelling or media, with a specific focus on dramatic, horror, and suspenseful narratives. &
    Categorization or analysis of music and entertainment genres, possibly including content recommendations or thematic associations.
    \\ \midrule

    \texttt{19-MLP-16K/ 5635}
    & References to political events and milestones. & 
    Concepts related to time measurement such as days, weeks, weekends, and months, indicating it likely pertains to scheduling or planning events. &
    Time periods, particularly weeks and weekends, along with some programming or markup elements for building or managing templates or components.
    \\ \midrule

    \texttt{21-RES-16K/ 10714}
    & Information related to bird species and wildlife activities. & 
    Concepts related to birdwatching and ornithology, focusing on activities such as observing, spotting, and recording bird species in their natural habitats. &
    Enhancing or analyzing bird watching or ornithological data and experiences, possibly improving the tracking of bird sightings and interactions.
    \\ \midrule

    \texttt{19-MLP-16K/ 1450}
    & Mentions of notable locations, organizations, or events, particularly in various contexts. &
    Concepts related to self-reflection, purpose, and generalization in various contexts, 
    focusing on the exploration of identity and overarching themes in literature or philosophy. &
    Recognize and generate variations of the term "general" and its context, along with concepts associated with insight and observation.
    \\ \bottomrule
\end{tabular}}
\caption{Example descriptions by \maxact{}, \vocabproj{} and \tokenshift{} for 4 SAE features from GemmaScope.}
\label{table:additional_human_eval}
\end{table*}

\section{Resources and Packages}
\label{apx:resources}

In our experiments, we used models, data and code from the following packages: transformers \cite{wolf2019huggingface}, datasets \cite{lhoest-etal-2021-datasets}, TransformerLens \cite{nanda2022transformerlens} and SAELens \cite{bloom2024saetrainingcodebase}. The authors also made use of AI models, specifically ChatGPT, for implementing specific helper functions. All of the experiments were conducted using a single A100 80GB or H100 80GB GPU.

\clearpage

\end{document}